\documentclass[11pt]{article}

\usepackage[final]{acl}

\usepackage{times}
\usepackage{latexsym}
\usepackage[T1]{fontenc}
\usepackage[utf8]{inputenc}
\usepackage{microtype}
\usepackage{inconsolata}
\usepackage{graphicx}

\usepackage{amsmath}
\usepackage{amsfonts}
\usepackage{nicefrac}
\usepackage{booktabs}
\usepackage{multirow}
\usepackage{float}
\usepackage{xcolor}
\usepackage{url}
\usepackage{tikz}
\usetikzlibrary{positioning, shapes.geometric, arrows.meta, fit, backgrounds}

\title{Retrieval-Augmented Generation Must Move Beyond Factual Grounding to Represent Diverse Opinions}

\author{
  Aditya Agrawal \quad Alwarappan Nakkiran \quad Alex Karlsson \quad Aman Singh Thakur \quad Harsha Aduri \\
  Amazon.com \\
  \texttt{\{aadag, nakkiran, nkarlss, amanzing, venaduri\}@amazon.com}
}

\begin{document}

\maketitle

\begin{abstract}
Retrieval-Augmented Generation (RAG) systems are built on an unexamined assumption: that queries have correct answers and retrieval should converge toward them. This position paper argues that this creates a factual bias where RAG systems optimize for reducing epistemic uncertainty while ignoring the aleatoric uncertainty, inherent in opinion-rich content. The consequences go beyond technical limitations: due to risk of minority voice erasure and risk of opinion manipulation. To address this, we formalize opinion-aware retrieval through uncertainty quantification and derive a unified objective using the Wasserstein distance. As an existence proof, we present Opinion-Aware RAG (O-RAG), which enriches documents with LLM-extracted, entity-linked opinion metadata before indexing. Across e-commerce seller forums and public hotel reviews, O-RAG reduces Wasserstein distance to corpus-level sentiment distributions by 18–48\%, and human evaluators preferred its responses 79.2\% of the time. We close with a research agenda for opinion-aware RAG.
\end{abstract}
\section{Introduction}

\begin{figure*}[t!]
\centering
\begin{tikzpicture}[
    node distance=0.6cm and 1.0cm,
    box/.style={rectangle, draw=black, thick, minimum height=0.7cm, minimum width=1.8cm, align=center, font=\scriptsize},
    enrichbox/.style={rectangle, draw=blue!70, thick, fill=blue!8, minimum height=0.8cm, minimum width=2.2cm, align=center, font=\scriptsize},
    attrbox/.style={rectangle, draw=gray!60, thick, fill=gray!5, minimum height=1.8cm, minimum width=2.0cm, align=left, font=\tiny},
    registrybox/.style={rectangle, draw=orange!70, thick, fill=orange!8, minimum height=0.55cm, minimum width=2.0cm, align=center, font=\scriptsize},
    arrow/.style={->, thick, >=stealth},
    dashedarrow/.style={->, thick, >=stealth, dashed, gray},
    label/.style={font=\tiny, text=gray!70}
]

\node[box] (raw) {Raw\\Documents};
\node[enrichbox, right=of raw] (enrich) {Opinion\\Enrichment\\(LLM-based)};
\node[box, right=of enrich] (enriched) {Enriched\\Documents};
\node[box, right=of enriched] (store) {Vector\\Store};
\node[box, right=of store] (llm) {LLM\\Generation};

\draw[arrow] (raw) -- (enrich);
\draw[arrow] (enrich) -- (enriched);
\draw[arrow] (enriched) -- (store);
\draw[arrow] (store) -- (llm);

\node[registrybox, above=0.3cm of enrich] (registry) {Entity Registry};
\node[label, right=0.1cm of registry] {(Tier 1 $\rightarrow$ 2 $\rightarrow$ 3)};
\draw[dashedarrow] (registry) -- (enrich);

\node[attrbox, below=0.9cm of enrich, xshift=0.4cm] (opattr) {
\textbf{Opinion Attributes}\\[1pt]
\textbullet~Sentiment\\
\textbullet~Intensity\\
\textbullet~Business Impact\\
\textbullet~Emotional Markers\\
\textbullet~Evidence Type
};

\node[attrbox, right=0.3cm of opattr] (authattr) {
\textbf{Author Attributes}\\[1pt]
\textbullet~Revenue Band\\
\textbullet~Tenure\\
\textbullet~Region\\
~\\
~
};

\draw[arrow, gray!60] (enrich.south) -- ++(0,-0.3) -| (opattr.north);
\draw[arrow, gray!60] (enrich.south) -- ++(0,-0.3) -| (authattr.north);

\node[below=3.6cm of raw, anchor=west, font=\scriptsize\itshape, text=gray!70] (tradlabel) {Traditional RAG:};
\node[box, right=0.2cm of tradlabel, fill=gray!10, draw=gray!50] (traw) {Raw\\Documents};
\node[right=0.4cm of traw, font=\normalsize, text=gray!50] (tarrow1) {$\rightarrow$};
\node[box, right=0.4cm of tarrow1, fill=gray!10, draw=gray!50] (tstore) {Vector\\Store};
\node[right=0.4cm of tstore, font=\normalsize, text=gray!50] (tarrow2) {$\rightarrow$};
\node[box, right=0.4cm of tarrow2, fill=gray!10, draw=gray!50] (tllm) {LLM};

\begin{scope}[on background layer]
\draw[draw=blue!30, thick, dashed, rounded corners=4pt, fill=blue!3]
    ([xshift=-6pt, yshift=6pt]enrich.north west |- registry.north)
    rectangle
    ([xshift=6pt, yshift=-6pt]enriched.south east |- authattr.south);
\end{scope}

\node[above=0.4cm of registry, font=\scriptsize\bfseries, text=blue!70] {Opinion Enrichment Step};

\end{tikzpicture}
\caption{Opinion-Aware RAG Architecture. Traditional RAG (bottom, grayed) indexes raw documents directly. Our approach introduces an opinion enrichment step (highlighted) that extracts per-entity opinion and author attributes before indexing. See Section~\ref{sec:architecture} for details.}
\label{fig:architecture}
\end{figure*}
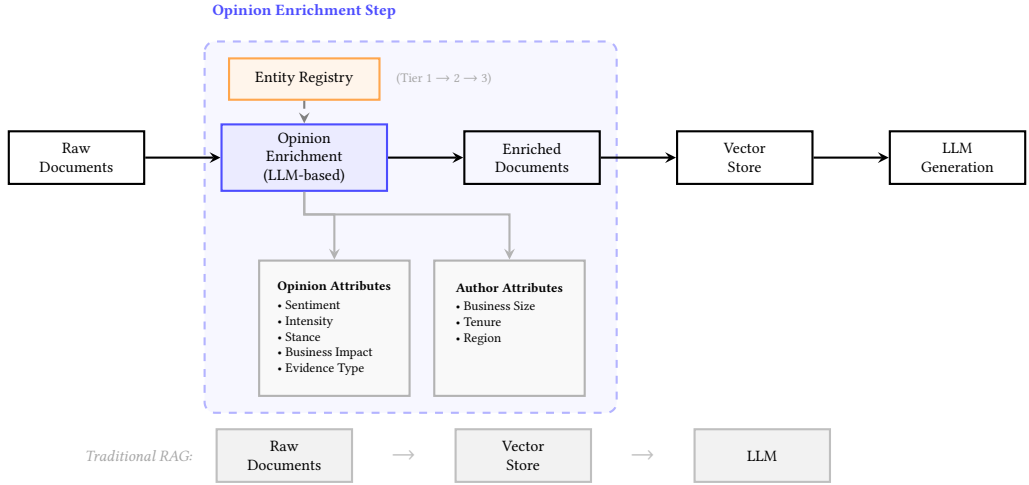

Retrieval-Augmented Generation has become a central technique for grounding Large Language Models in external evidence, reducing hallucinations and enabling verifiable, up-to-date text generation~\citep{lewis2020retrieval, gao2023retrieval, gan2024retrieval}. However, beneath this progress lies an unexamined assumption- that knowledge bases should contain factually correct, objective information, and that retrieval would converge toward the ground-truth. This assumption pervades the RAG ecosystem, from the datasets used for evaluation, to the retrieval mechanisms optimized for factual recall, to the generation objectives that reward consistency with a single correct answer.

\begin{table}[t!]
\caption{Representative RAG evaluation datasets (9 of 36 surveyed). Only OpinioRAG addresses opinion synthesis; PIR and BERDS evaluate perspective-aware retrieval but not distributional fidelity. See Appendix~\ref{app:benchmarks} for the complete analysis.}
\label{tab:rag_benchmarks}
\small
\centering
\setlength{\tabcolsep}{4pt}
\begin{tabular}{lclc}
\toprule
Dataset & Year & Evaluation Focus & Opinion \\
\midrule
OpinioRAG & 2025 & Sentiment Alignment & Yes \\
BERDS & 2025 & Perspective Coverage & Partial \\
PIR & 2024 & Perspective Recall & Partial \\
RAGBench & 2024 & TRACe Metrics & No \\
CRAG & 2024 & Accuracy, Hallucination & No \\
FActScore & 2023 & Atomic Fact Accuracy & No \\
BEIR & 2021 & nDCG, MAP, Recall & No \\
Natural Questions & 2019 & EM, Token F1 & No \\
HotpotQA & 2018 & EM, F1, Sup.\ F1 & No \\
\bottomrule
\end{tabular}
\end{table}

Of 36 major RAG benchmarks surveyed (Table~\ref{tab:rag_benchmarks}), only one addresses opinion synthesis- OpinioRAG~\citep{nayeem2025opiniorag}. Two others (PIR~\citep{zhao2024pir}, BERDS~\citep{chen2025berds}) evaluate perspective-aware retrieval but not distributional fidelity. The rest assume questions have definitive answers and treat contradictory information as noise. This reveals a misalignment between how RAG systems are built and how a large share of real-world information actually behaves.

Consider two queries: \textit{``What is the corporate tax rate in the United States?''} versus \textit{``What do small business owners think about the current tax policy?''} The first has a definitive answer that existing RAG systems handle well. The second requires synthesizing diverse, potentially conflicting perspectives from multiple sources. This is a task for which current architectures have no explicit design. Some LLMs demonstrate emergent opinion synthesis capabilities- producing balanced-sounding summaries that capture diverse viewpoints (see Appendix~\ref{app:responses} for an example). However, the difference between accidental capability and intentional design matters when deploying systems, where understanding the sources and potential biases of opinion synthesis is critical.

\textbf{We argue that current RAG systems exhibit a factual bias. They optimize for epistemic uncertainty reduction, while ignoring the aleatoric uncertainty inherent in opinion-rich content.}

This limitation grows more consequential as RAG deployments expand beyond traditional knowledge bases into online community forums, consumer reviews, social media, and user-generated content~\citep{rony2025benchmark, chen2025supply}. In these domains, the goal is to understand and represent \textit{multiple truths} as expressed through human experiences and judgments, instead of converging on single truth. Without opinion-aware capabilities, systems risk echo chamber amplification, minority voice erasure, potential for opinion manipulation, and artificial consensus where genuine disagreement exists. While recent work on pluralistic alignment~\citep{sorensen2024value, bakker2022finetuning} addresses diversity at the generation layer, it cannot compensate for biased retrieval, as a generation model can only reason over what it receives.

Our contributions are:
\begin{enumerate}
    \item An \textbf{audit} of factual bias across the RAG ecosystem. We document the absence of opinion-aware evaluation (Section~\ref{sec:audit}).
    \item A \textbf{theoretical framework} formalizing opinion-aware retrieval through epistemic vs.\ aleatoric uncertainty (Section~\ref{sec:theory}).
    \item An \textbf{existence proof}: Opinion-Aware RAG evaluated across two domains (e-commerce forums and public hotel reviews) using a novel evaluation framework measuring opinion coverage, perspective diversity, and representation fairness. We demonstrate that even simple opinion enrichment yields measurably more representative retrieval and generation (Sections~\ref{sec:architecture}--\ref{sec:experiments}, Figure \ref{fig:architecture}).
    \item A \textbf{research agenda} identifying open problems for the community (Section~\ref{sec:agenda}).
\end{enumerate}

Crucially, our position is a claim about \textit{generation}, not retrieval alone: the objective is opinion-preserving text generation, where the distribution of perspectives is faithfully carried through to the produced output, since a system that surfaces diverse viewpoints but collapses them into an artificial consensus during generation still fails the user.

\section{The Factual Bias in RAG: An Audit}
\label{sec:audit}

The factual bias in RAG is a field-wide pattern that runs through the entire pipeline, from how benchmarks are constructed to how retrieval and evaluation are designed. We document this through a survey of 36 major RAG evaluation datasets (2013--2025), summarized in Table~\ref{tab:rag_benchmarks} and detailed in Appendix~\ref{app:benchmarks}.

The pattern is consistent across task categories: single-hop QA (Natural Questions, MS~MARCO, SQuAD), multi-hop reasoning (HotpotQA, MuSiQue), long-form QA (ELI5, ASQA), domain-specific QA (COVID-QA, Qasper), fact verification (FEVER, TruthfulQA), and generation quality (FActScore, RAGBench). Every category assumes questions have correct answers and evaluates systems on their ability to converge toward them.

This bias becomes consequential as RAG deployments expand into domains where understanding human opinion is the primary goal~\citep{zhao2024retrieval, wang2025systematic, fan2024survey}: online community forums, consumer reviews, customer service, policy analysis, healthcare patient feedback, and political discourse. In these settings, information is inherently subjective, and collapsing it to a single answer misrepresents reality. Yet recent work actively reinforces this mismatch by framing subjectivity as noise: RAGuard~\citep{zeng2025worse} treats Reddit discussions as naturally occurring misinformation to be filtered, CONFLICTS~\citep{cattan2025dragged} focuses on conflict resolution rather than preserving perspective diversity, and robustness approaches~\citep{chen2022rich, choi2025conflict, zhang2025faithfulrag} emphasize filtering contradictory information. While valuable for factual domains, these approaches are misaligned with scenarios where diverse perspectives are the goal.

The boundary between factual and opinion content is blurred in practice. Even a factual query, such as ``What are the shipping rates for small parcels?'', will surface opinion-laden forum discussions alongside authoritative sources when the knowledge base includes community content. A semantic similarity-based retriever cannot distinguish an official rate schedule from a seller's complaint about rate increases. Current systems have no mechanism to handle this mixture, treating all retrieved content as equivalent evidence. Existing diversity methods do not solve this: Maximal Marginal Relevance~\citep{carbonell1998mmr} and Determinantal Point Process-based methods promote diversity by penalizing redundancy in embedding space, but semantic diversity and opinion diversity are fundamentally different objectives. Two documents expressing opposing views on the same topic may use similar vocabulary and land close together in embedding space. MMR would penalize retrieving both; opinion-aware retrieval should specifically surface both.

Additionally, current RAG-based metrics can assign high scores to biased outputs. A response to \textit{``What leadership qualities are most important for startup CEOs?''} that emphasizes only traditionally masculine-coded traits could score excellently on ROUGE-L, context relevance, and answer faithfulness. However, it would underrepresent collaborative and inclusive leadership perspectives~\citep{wang2025bias, zhang2025evaluating, oliveira2025fairness, thakur2025judging} (see Appendix~\ref{app:biased_leadership} for a concrete example). Traditional metrics tend to reward convergence and cannot detect what was excluded.

Related techniques like argument mining and stance detection have matured considerably~\citep{aldayel2021stance, zhang2024survey}, with work spanning argument-based retrieval~\citep{li2025argument}, sentiment analysis on decentralized platforms~\citep{chhetri2025cognitiveskybluesky}, and multimodal/comment-level opinion analysis~\citep{sharma2020memotion, li2018youtube, weissmann2025youtube, sun2024japanese}, yet these capabilities remain isolated from RAG frameworks. RAFTS~\citep{yue2024retrieval} synthesizes contrastive arguments but targets claim credibility, not opinion diversity. LLMs' inability to reliably detect contradictions~\citep{lee2024magic, gokul2025contradiction} compounds the challenge. OpinioRAG~\citep{nayeem2025opiniorag} bridges this gap via RAG-based opinion summarization but focuses on condensing opinions into narratives rather than ensuring the retrieved set faithfully represents the opinion spectrum.

Latest work on perspective benchmarks is a step in this direction, but is incomplete. PIR~\citep{zhao2024pir} and BERDS~\citep{chen2025berds} confirm that current retrievers are perspective-blind: PIR shows systematic bias toward supporting stances (${\sim}$58.5\%), while BERDS finds full perspective coverage in only 40\% of cases. However, both evaluate categorical \textit{perspective} presence (is viewpoint X retrieved?) rather than \textit{opinion} distributions - the intensity, affect, and demographic grounding that characterize real-world subjective content.

Pluralistic alignment is another relevant body of work which is complementary to opinion-aware retrieval. A growing body of work addresses pluralism at the generation layer: \citet{sorensen2024value} formalize pluralistic value alignment for LLMs, \citet{bakker2022finetuning} fine-tune models to find agreement among diverse preferences, and \citet{argyle2023silicon} frame LLMs as ``silicon sampling'' of human opinions. However, pluralistic alignment trains the \textit{generation model} to preserve opinion distributions, while opinion-aware retrieval ensures the \textit{retrieved context} faithfully represents the distribution \textit{before} generation. These are complementary: a pluralistically-aligned LLM given biased retrieved context still produces biased responses. Moreover, without transparency into retrieval, users cannot distinguish substantive balance from stylistic tendencies induced by RLHF~\citep{sharma2023sycophancy}. Even when diverse opinions are retrieved, users cannot understand how conflicting perspectives were weighted or what constitutes ``balanced representation''. This is different from factual retrieval where citation suffices, as opinion synthesis requires understanding how conflicting viewpoints are reconciled.

\section{Epistemic vs.\ Aleatoric: A Theoretical Foundation}
\label{sec:theory}

We formalize the distinction between factual and opinion-aware retrieval through uncertainty quantification. Throughout, we use: $q$ (query), $Y$ (the response, with realization $y$), $\mathcal{D}$ (corpus), $\theta$ (latent state), $P_{\text{pop}}(\theta)$ (population opinion distribution), $\mathcal{R}(q)$ (retrieved documents), $\phi: \mathcal{D} \to \mathbb{R}^m$ (opinion embedding), $H(\cdot)$ (entropy), and $\mathcal{W}_p$ ($p$-Wasserstein distance).

\subsection{Two Kinds of Uncertainty}

The posterior predictive distribution for a query $q$ given corpus $\mathcal{D}$ is:
\begin{equation}
P(y|q, \mathcal{D}) = \int P(y|\theta, q) \, P(\theta|\mathcal{D}) \, d\theta
\end{equation}

For \textbf{factual queries}, a unique true state $\theta^*$ exists. Uncertainty is \textit{epistemic} and reducible through evidence. As corpus size grows, the posterior collapses to a point:
\begin{equation}
\lim_{|\mathcal{D}| \to \infty} P(\theta|\mathcal{D}) = \delta(\theta - \theta^*)
\end{equation}

For \textbf{opinion queries}, uncertainty is \textit{aleatoric} and reflects genuine heterogeneity rather than ignorance. The posterior converges to the population distribution, not a point:
\begin{equation}
\lim_{|\mathcal{D}| \to \infty} P(\theta|\mathcal{D}) = P_{\text{pop}}(\theta)
\end{equation}

This distinction has a direct design implication: \textit{factual RAG should minimize posterior entropy; opinion-aware RAG must preserve it.} We note that opinion queries still involve epistemic uncertainty about $P_{\text{pop}}$ itself. However, the goal is to reduce this epistemic component through comprehensive retrieval, while preserving the aleatoric component by representing discovered diversity.

\subsection{Divergent Optimization Objectives}

Standard RAG maximizes information gain by minimizing conditional entropy:
\begin{equation}
\mathcal{L}_{\text{factual}} = H(Y|q) - H(Y|q, \mathcal{R}(q))
\end{equation}
This rewards retrieval that eliminates uncertainty which works correctly for facts and is counterproductive for opinions.

For opinions, we seek \textit{distributional fidelity}. Let $v$ denote a position in opinion space and $P_{\text{pop}}(v|q)$ the population opinion distribution. The opinion-aware objective minimizes divergence while penalizing artificial consensus:
\begin{equation}
\mathcal{L}_{\text{OA}} = \mathcal{W}_2\!\left( P_{\text{pop}}(v|q),\; P_{\text{model}}(v|q) \right) - \lambda \, H(P_{\text{model}})
\end{equation}

\subsection{Retrieval as Coverage Optimization}

Let $\mathcal{R}(q) = \{d_1, \ldots, d_k\}$. The embedding $\phi: \mathcal{D} \to \mathbb{R}^m$ maps documents to opinion space; in our implementation (Section~\ref{sec:architecture}), $\phi$ encodes sentiment polarity, stance intensity, and topical framing. The empirical opinion distribution from retrieval is:
\begin{equation}
\hat{P}_{\mathcal{R}} = \frac{1}{k} \sum_{i=1}^{k} \delta_{\phi(d_i)}
\end{equation}

We measure coverage using the $p$-Wasserstein distance, where $\Gamma(\cdot,\cdot)$ denotes the set of couplings between the two distributions:
\begin{equation}
\mathcal{W}_p(\hat{P}_{\mathcal{R}}, P_{\text{pop}}) = \left( \inf_{\gamma \in \Gamma} \int \|\phi - \phi'\|^p \, d\gamma \right)^{1/p}
\end{equation}
Intuitively, $\mathcal{W}_p$ measures the minimum cost to reshape one distribution into another, naturally penalizing retrieval that misses regions of opinion space. We prefer it over KL divergence or Jensen-Shannon divergence because it remains well-defined when distributions have non-overlapping support.

\subsection{Unified Objective}

Opinion-Aware RAG minimizes:
\begin{equation}
\label{eq:unified}
\begin{aligned}
\mathcal{L} = {}&\underbrace{\mathcal{W}_2(P_{\text{pop}}, \hat{P}_{\mathcal{R}})}_{\text{coverage}} + \alpha \underbrace{\mathcal{W}_2(P_{\text{pop}}, P_{\text{reader}})}_{\text{fidelity}} \\
&+ \beta \underbrace{\max_{g \in G} \Delta_g}_{\text{fairness}}
\end{aligned}
\end{equation}
where $P_{\text{reader}}(\phi | y)$ is the opinion distribution a reader infers from response $y$, and $\Delta_g = \mathcal{W}_2(P_{\text{pop}}(\phi | g), \hat{P}_{\mathcal{R}}(\phi | g))$ measures representation disparity for demographic group $g \in G$. The minimax fairness criterion ensures no subpopulation is excluded ($\alpha, \beta > 0$).

\paragraph{Scope and limitations.} This framework identifies \textit{what} opinion-aware RAG should optimize - coverage, fidelity, and fairness. However, this is not a directly implementable loss function. Key assumptions include: (1)~$P_{\text{pop}}$ is estimated from the corpus, treating it as a population sample; (2)~opinions are treated as static (temporal dynamics are out of scope for this work); (3)~directly optimizing Equation~\ref{eq:unified} is computationally prohibitive with current retrieval architectures. Our implementation (Section~\ref{sec:architecture}) approximates coverage through opinion-enriched indexing, and we validate the connection empirically via Wasserstein distance measurements in Section~\ref{sec:experiments}.

This formulation frames opinion-aware RAG as a \textit{distribution estimation and communication problem}. This framing differs substantially from the point-estimation paradigm underlying factual retrieval, and one that we believe opens productive directions for future work.

\section{O-RAG: A Practical Architecture}
\label{sec:architecture}

We present Opinion-Aware RAG (O-RAG) as one possible instantiation of the theoretical framework. We do not claim this to be the definitive solution, but as evidence that even simple opinion-aware enrichment yields measurable improvements. The architecture introduces an \textbf{opinion enrichment step} before indexing (illustrated in Figure~\ref{fig:architecture}), consisting of three domain-agnostic components.

\paragraph{Entity registry.} A hierarchical taxonomy of opinion targets organized by granularity. Entities represent the topics people hold views about. The registry can be bootstrapped automatically via LLM extraction on sampled documents, followed by clustering and deduplication. Manual curation helps refine but is not strictly required.

\paragraph{Opinion attributes.} Structured metadata capturing the nature of expressed opinions: sentiment and intensity, perceived impact or severity, emotional markers, and evidence type (personal experience, anecdotal, or data-driven).

\paragraph{Author attributes.} Contextual factors about opinion holders that enable fairness analysis. These can be demographic or experiential characteristics relevant to the domain, supporting the fairness term $\max_{g \in G} \Delta_g$ in the unified objective (Equation~\ref{eq:unified}).

\paragraph{Per-entity document splitting.} Each source document is split into multiple enriched documents (one per mentioned entity) with opinion and author attributes embedded alongside content. This ensures that when a document discusses multiple topics with different sentiments, each opinion is indexed separately with its corresponding metadata.

\subsection{Instantiation}

We instantiate this framework in two domains:

\textbf{E-commerce seller forums.} Sellers discuss policies, fees, pain points, and operational challenges. We construct a 98-entity registry across three tiers, extract four opinion attributes (sentiment, intensity, business impact, emotional markers) and three author attributes (revenue band, tenure, region) per entity mention using an LLM with structured output (see Appendix~\ref{app:extraction_prompt} for the full extraction prompt).

\textbf{Public hotel reviews (OpinioBank).} Consumer reviews of hotel properties, drawn from the publicly available OpinioBank dataset introduced by \citet{nayeem2025opiniorag}. The extraction prompt required only minor domain adaptation; entity registry construction stabilized at approximately 90~entities. This demonstrates cross-domain feasibility with minimal manual effort.

Both configurations use vector search with Amazon Titan Text Embeddings~V2 (1024 dimensions) and fixed-size chunking (4000 tokens, 15\% overlap). We compare an Opinion-Enriched KB against a Raw KB (traditional RAG with no opinion metadata) using identical source content. Full model, decoding, retrieval, and cost settings are reported in Appendix~\ref{app:reproducibility}.

\section{Empirical Evidence Across Domains}
\label{sec:experiments}

These experiments are not meant to prove O-RAG is the optimal solution, but to demonstrate that (a)~the problem is real and measurable, and (b)~even simple opinion-aware enrichment yields significant improvements. We evaluate across two domains spanning 19 diverse entities and 105 manually curated questions.

\subsection{Setup}

\paragraph{Datasets.} (1)~\textit{E-commerce seller forums}: six months of discussions, approximately 8,000 unique sellers, 10K+ documents. (2)~\textit{OpinioBank hotel reviews}~\citep{nayeem2025opiniorag}: we use a sample of three hotel properties (Silverton Las Vegas, Ambassador Bangkok, Ala Moana Honolulu) from the full 500-entity OpinioBank release, totaling 6,256~reviews across 15~entities and 57~questions.

\paragraph{Evaluation protocol.} We target entities with genuine opinion diversity by computing sentiment entropy $H(e)$ and selecting those with $H(e) \geq 0.6$ ensuring both positive and negative sentiments are represented (4 entities for forums, 15 for OpinioBank). For each entity, we generate natural language questions across multiple types designed to probe different facets of opinion. This yielded 48~questions for forums and 57 for OpinioBank. Each question is submitted to both KB configurations. Then, we retrieve the top-$k$ documents and generate responses using the retrieved context. Retrieved documents are matched back to opinion metadata to compute diversity metrics.

\paragraph{Metrics.} We design proxy metrics aligned with the theoretical objectives (Equation~\ref{eq:unified}). 

For \textit{coverage}: (1)~\textbf{Sentiment-Intensity $\sigma$}: combines sentiment polarity and intensity into a single numeric score (Positive+High = +30, Positive+Medium = +20, Positive+Low = +10, Neutral = 0, Negative+Low = $-$10, Negative+Medium = $-$20, Negative+High = $-$30); we report the standard deviation across retrieved documents, where higher $\sigma$ indicates broader coverage of the sentiment-intensity space. (2)~\textbf{Business Impact $\sigma$}: maps impact levels to numeric values (High = 30, Medium = 20, Low = 10); higher $\sigma$ indicates retrieval of both high-impact and low-impact perspectives. (3)~\textbf{Entity Match Rate}: fraction of retrieved documents that mention the queried entity, measuring retrieval precision for opinion-relevant content. (4)~\textbf{Emotional Markers}: count of distinct emotional markers (e.g., Frustration, Hope, Resignation) across retrieved documents. (5)~\textbf{Empirical $\mathcal{W}_1$}: Wasserstein distance between corpus-level and retrieved sentiment distributions. 

For \textit{fidelity}: human evaluation via blind pairwise comparison on the forums dataset (5~annotators, 48~pairs). 

For \textit{fairness}: \textbf{Demographic Coverage}, the fraction of categories represented in retrieved documents for each author attribute: Revenue Band (business size), Tenure (experience levels), and Evidence Type (personal, anecdotal, data-driven).

\subsection{Results}

\paragraph{Coverage.} Table~\ref{tab:main_results} presents the core comparison at $k{=}20$ with hybrid retrieval on the forums dataset. Opinion enrichment significantly improves all coverage metrics: +42.7\% entity match rate, +26.8\% sentiment-intensity $\sigma$, +14.5\% business impact $\sigma$, and +23.6\% emotional markers (all $p{<}0.05$).

\begin{table}[t]
\caption{Diversity metrics at $k{=}20$ (hybrid retrieval, forums dataset). Statistical significance via Wilcoxon signed-rank test ($n{=}48$): \textbf{bold}~=~$p{<}0.05$, *~=~$p{<}0.001$.}
\label{tab:main_results}
\small
\centering
\setlength{\tabcolsep}{4pt}
\begin{tabular}{lccc}
\toprule
\textbf{Metric} & \textbf{Raw KB} & \textbf{Enr.\ KB} & $\boldsymbol{\Delta}$ \\
\midrule
Entity Match (\%) & 24.8 & 35.4 & \textbf{+42.7\%}* \\
Sent-Int $\sigma$ & 9.12 & 11.56 & \textbf{+26.8\%}* \\
Biz Impact $\sigma$ & 10.55 & 12.08 & \textbf{+14.5\%} \\
Emot.\ Markers & 7.2 & 8.9 & \textbf{+23.6\%}* \\
\midrule
Revenue Band (\%) & 78.3 & 68.1 & \textbf{$-$13.0\%}* \\
Tenure (\%) & 81.2 & 75.6 & \textbf{$-$6.9\%} \\
Evidence (\%) & 44.3 & 46.4 & +4.7\% \\
\bottomrule
\end{tabular}
\end{table}

To directly validate the connection to our theoretical framework, we compute empirical $\mathcal{W}_1$ between corpus-level and retrieved sentiment distributions (Table~\ref{tab:wasserstein}). The enriched KB is consistently 18-48\% closer to corpus-level sentiment truth across both domains.

\begin{table}[t]
\caption{Empirical $\mathcal{W}_1$ between corpus-level and retrieved sentiment distributions ($k{=}20$). Lower is better. Forums results significant at $k{=}10, 40, 60$ ($p{<}0.05$).}
\label{tab:wasserstein}
\small
\centering
\begin{tabular}{lccc}
\toprule
\textbf{Dataset} & \textbf{Raw $\mathcal{W}_1$} & \textbf{Enriched $\mathcal{W}_1$} & $\boldsymbol{\Delta}$ \\
\midrule
Forums & 8.87 & 7.24 & $-$18\% \\
Silverton & 15.31 & 9.17 & $-$40\% \\
Ambassador & 19.15 & 14.54 & $-$24\% \\
Ala Moana & 12.88 & 6.75 & $-$48\% \\
\bottomrule
\end{tabular}
\end{table}

\paragraph{Fidelity.} Human annotators confirmed that opinion-enriched retrieval produces responses that better communicate opinion diversity. Of 48 question-response pairs, annotators preferred Opinion-Enriched KB responses 79.2\% of the time (95\% CI: [65.0\%, 89.5\%], $p{<}0.001$). Appendix~\ref{app:annotation} has more details about the annotation study. Appendix~\ref{app:qualitative} provides a qualitative example showing how enriched responses capture sentiment distributions and constructive suggestions missing from baseline responses.

\begin{table*}[t]
\caption{OpinioBank results (public hotel reviews). Entity match improvement and sentiment-intensity $\sigma$ change vs.\ Raw KB.}
\label{tab:opiniobank}
\small
\centering
\begin{tabular}{lcccc}
\toprule
\textbf{Property} & \textbf{Reviews} & \textbf{Entities} & \textbf{Entity Match $\Delta$} & \textbf{Sent-Int $\sigma$ $\Delta$} \\
\midrule
Silverton LV & 2,297 & 4 & +37.9pp & +5.04 \\
Ambassador Bangkok & 1,230 & 4 & +37.2pp & +7.09 \\
Ala Moana Honolulu & 2,729 & 4 & +18.9pp & +3.00 \\
\bottomrule
\end{tabular}
\end{table*}

\paragraph{Fairness.} Contrary to expectations, opinion enrichment reduces overall demographic coverage: Revenue Band by 13.0\% and tenure by 6.9\% (both $p{<}0.05$). However, when comparing only documents that actually mention the queried entity, demographic coverage \textit{improves} (Revenue Band +9.6\%, Tenure +31.6\%; see Appendix~\ref{app:entity_matched}). This reveals that the Raw KB's apparent demographic breadth comes from off-topic documents. It retrieves from a wider author pool, but much of that content is irrelevant to the query. The Enriched KB concentrates retrieval on entity-relevant content, achieving better demographic representation among documents that matter. This trade-off motivates future work on multi-objective retrieval.

\paragraph{Cross-domain generalization.} Table~\ref{tab:opiniobank} shows consistent improvements on public OpinioBank data across a different domain (hotel reviews vs.\ e-commerce forums). Diversity metrics were statistically significant for emotional markers; direction and magnitude are consistent with the forums findings. On OpinioBank, entity-matched documents show consistent demographic improvements (+12.5 to +29.4pp), confirming the fairness pattern observed on forums.

\subsection{Ablations}

\paragraph{Cross-model consistency.} The Full O-RAG improvement over Raw KB is consistent across extraction models: Mistral Large (+24.2\% Sent-Int $\sigma$), Qwen 235B (+27.4\%), and Claude Sonnet 4.5 (+26.8\%), confirming the approach is not dependent on a specific LLM (detailed in Appendix~\ref{app:model_variations}).

\paragraph{Entity splitting vs.\ full O-RAG.} An ablation disentangling per-entity splitting from opinion metadata reveals a nuanced picture (see Appendix~\ref{app:split_ablation} for full tables). Both Entity-Split-Only and Full O-RAG substantially improve over Raw KB across all metrics. For Mistral Large, Full O-RAG outperforms Entity-Split-Only on all diversity metrics ($p{<}0.001$): Sent-Int $\sigma$ 8.11 vs.\ 7.37, Biz Impact 10.94 vs.\ 10.27, Emot Markers 4.1 vs.\ 3.5. For Sonnet 4.5, Entity-Split-Only achieves higher Sent-Int $\sigma$ (13.1 vs.\ 11.56) and Biz Impact (12.92 vs.\ 12.08), but Full O-RAG wins on Emotional Markers (8.9 vs.\ 7.8, $p{<}0.05$). This suggests that entity splitting captures a large share of diversity gains through structural reorganization, while opinion metadata contributes additional value (particularly for emotional and affective dimensions) that varies by extraction model.

\paragraph{Question type variation.} Business Impact (+38.9\%) and Emotional (+34.1\%) queries benefit most from opinion enrichment, while Experience queries (+6.2\%) benefit least- likely because they already surface diverse personal perspectives without metadata (Appendix~\ref{app:question_type}).

\paragraph{Retrieval method.} Hybrid retrieval amplifies enrichment benefits over semantic-only search (+26.8\% vs.\ +21.4\% Sent-Int $\sigma$), as keyword matching surfaces entity-relevant documents (Appendix~\ref{app:retrieval_method}).

\paragraph{Retrieval depth.} Sent-Int $\sigma$ was consistently higher for opinion-enriched KB across varying retrieval depths (10-60) showing the performance is retrieval depth agnostic. (Appendix~\ref{app:depth})

\subsection{Alternative Views and Counterarguments}
\label{sec:counterarguments}

One might argue that the LLM handles opinion synthesis at generation time: that RAG retrieval is merely context and a capable LLM can synthesize diverse opinions from whatever it receives. Our experiments suggest otherwise. Raw KB retrieval systematically excludes perspectives before the LLM sees them: at $k{=}20$, the Raw KB's sentiment distribution is 18-48\% further from corpus truth than the Enriched KB's (Table~\ref{tab:wasserstein}). The LLM can only reason over what it receives. Human evaluators preferred enriched responses 79.2\% of the time ($p{<}0.001$), indicating that retrieval-level bias propagates through generation despite the LLM's synthesis capabilities.

A second objection could be that existing diversity methods and pluralistic alignment already solve this. Diversity-aware retrieval methods like MMR~\citep{carbonell1998mmr} promote diversity in embedding space, and pluralistic alignment~\citep{sorensen2024value, bakker2022finetuning} trains generation models to produce balanced outputs. As discussed in Section~\ref{sec:audit}, these address different dimensions of the problem. MMR diversifies based on semantic dissimilarity- but opposing opinions on the same topic can be semantically \textit{similar}, sharing vocabulary while differing in stance. MMR would penalize retrieving both. Pluralistic alignment operates at the generation layer and is complementary- a pluralistically-aligned LLM given biased retrieved context still produces biased responses.

A third objection is that opinion-aware RAG risks amplifying fringe or harmful viewpoints. However, the goal is \textit{transparent representation proportional to prevalence}, not equal amplification. The distributional fidelity objective (Equation~\ref{eq:unified}) weights opinions by corpus frequency, and the fairness term penalizes over- or under-representation of demographic groups. Opaque filtering carries its own risk: users cannot audit what was excluded, making bias invisible rather than absent.

\section{A Research Agenda for Opinion-Aware RAG}
\label{sec:agenda}

We identify seven open problems for the community:

\textbf{1.\ Opinion-aware benchmarks.} The field needs RAG evaluation datasets with naturally occurring subjective content and annotations for stance, sentiment, demographic attributes, and perspective diversity. Without such benchmarks, progress on opinion-aware retrieval remains difficult to measure.

\textbf{2.\ Learned opinion embeddings.} Our approach relies on LLM-based extraction, but the theoretical framework's opinion embedding $\phi: \mathcal{D} \to \mathbb{R}^m$ suggests the need for learned representations that jointly encode semantic content with subjective attributes. 

\textbf{3.\ Direct distributional optimization.} Our proxy metrics approximated the coverage objective $\mathcal{W}_2(P_{\text{pop}}, \hat{P}_{\mathcal{R}})$ but to effectively close the distributional gap we need retrieval strategies that directly minimize this during search - via query expansion, stance-diverse re-ranking, or submodular optimization.

\textbf{4.\ Multi-objective retrieval.} Our experiments reveal trade-offs between opinion diversity and demographic fairness. Retrieval strategies that jointly optimize the three terms in Equation~\ref{eq:unified} remain an open challenge.

\textbf{5.\ Temporal opinion dynamics.} The framework assumes a static $P_{\text{pop}}$, but opinions evolve. Sometimes, rapidly in response to events (e.g., a policy change triggering immediate backlash that later normalizes) and sometimes gradually as communities shift attitudes over months or years. Distinguishing temporary reactive opinions from lasting attitudinal shifts and detecting when $P_{\text{pop}}$ has changed enough to invalidate cached representations require extending the theoretical foundation to handle non-stationary distributions. This is particularly important for domains like e-commerce and social media where opinion landscapes can shift within days.

\textbf{6.\ Generation fidelity.} Our evaluation focuses on retrieval diversity. Measuring whether generated responses faithfully communicate retrieved opinion distributions to readers, the fidelity term $\mathcal{W}_2(P_{\text{pop}}, P_{\text{reader}})$, requires new evaluation protocols beyond pairwise preference.

\textbf{7.\ Joint retrieval-generation optimization.} Combining opinion-aware retrieval with pluralistic alignment at the generation layer could yield systems that are both representative in what they retrieve and faithful in what they communicate.

\section{Conclusion}

We have argued that RAG systems exhibit a  factual bias which is embedded throughout the generation pipeline from retrieval objectives to evaluation metrics, rendering misalignment with opinion-rich content. Our theoretical framework formalizes this through the epistemic--aleatoric distinction. We propose that factual queries should minimize posterior entropy, while opinion queries must preserve it. The unified objective over coverage, fidelity, and fairness provides a principled foundation for future work.

As an existence proof, Opinion-Aware RAG demonstrates that even simple opinion enrichment yields 18-48\% reduction in distributional divergence from corpus truth across two domains. Human evaluators confirm the perceptible quality improvements transfers to generated responses.

The path forward requires the community to treat opinion-aware RAG as a first-class research problem: building benchmarks with subjective annotations, learning opinion-aware representations, developing retrieval and generation strategies that jointly optimize for diversity, and establishing evaluation protocols for opinion synthesis. As AI systems increasingly mediate access to information, we argue that their ability to represent that diversity in the generated responses is critical.

\newpage
\section*{Limitations}

Our work has several limitations that also frame directions for future research.

\paragraph{Theoretical framework is not directly implementable.} The unified objective in Equation~\ref{eq:unified} identifies \textit{what} opinion-aware RAG should optimize (coverage, fidelity, and fairness), but it is not a directly optimizable loss under current retrieval architectures. It relies on three assumptions: (1)~the population opinion distribution $P_{\text{pop}}$ is estimated from the corpus and treated as a population sample, which may not hold when the corpus itself is a biased sample of the underlying population; (2)~opinions are treated as static, so temporal dynamics are out of scope; and (3)~directly minimizing the objective is computationally prohibitive. Our implementation approximates only the coverage term through opinion-enriched indexing.

\paragraph{Limited domain and scale of evaluation.} We evaluate on two English-language domains (e-commerce seller forums and hotel reviews) with 105 curated questions and 19 high-entropy entities. This is sufficient as an existence proof but does not establish generality across languages, cultures, or higher-stakes domains such as healthcare or political discourse. Entity selection was restricted to those with sentiment entropy $H(e) \geq 0.6$, which deliberately targets opinion-diverse content and may overstate gains relative to a uniformly sampled query set.

\paragraph{Human evaluation scope.} The fidelity evaluation rests on blind pairwise preference from 5 annotators over 48 question-response pairs on a single domain. The wide confidence interval (95\% CI: [65.0\%, 89.5\%]) reflects this limited scale, and we did not measure whether readers correctly \textit{infer} the underlying opinion distribution (the fidelity term $\mathcal{W}_2(P_{\text{pop}}, P_{\text{reader}})$).

\paragraph{Dependence on LLM-based extraction.} Opinion attributes and the entity registry are produced by LLM extraction. While we show cross-model consistency across three extraction models, the extraction step can itself introduce a systematic bias (for example, in how sentiment intensity or demographic attributes are assigned), and we do not fully audit these downstream effects.

\paragraph{Diversity-fairness trade-off.} Opinion enrichment reduces overall demographic coverage even as it improves coverage among entity-relevant documents. We characterize this trade-off empirically but do not resolve it; multi-objective retrieval that jointly optimizes diversity and fairness remains open.

\section*{Ethics Statement}

This work is motivated by an ethical concern: RAG systems that collapse opinion-rich content to a single answer can lead to echo chamber effects that amplify dominant viewpoints, which can lead to opinion manipulation and under-representation of minority voices. Our aim is transparent representation in proportion to prevalence rather than equal dissemination of all views. We acknowledge a dual-use tension. A system that faithfully represents an opinion distribution could also be used to surface fringe or harmful viewpoints. We argue that the distributional-fidelity and fairness terms in our objective mitigate this by weighting opinions according to corpus prevalence. The objective penalizes over or under-representation of demographic groups and that transparent representation is preferable to opaque filtering that hides, what was excluded. We do not claim this fully resolves the risk, and deployment in high-stakes settings would require additional safeguards.

\paragraph{Human annotation.} All annotation was performed by internal colleagues (middle-aged, US-based, fluent English speakers) as part of their normal work; no additional compensation was provided for the task. Five evaluators completed the blind pairwise preference comparisons, and three annotators produced the opinion labels used to measure inter-annotator agreement. The annotators consented to the use of their judgments for this research. The study collected no personal data beyond the annotators' preference and label judgments over publicly derived text and was determined not to require ethics-board review. Annotators reviewed user-generated content that may contain complaints or negative sentiment but no sensitive personal information beyond what the source platforms already make available. No demographic inferences about individuals were released.

\paragraph{Inferred author attributes.} The architecture derives author attributes (for example, revenue band, tenure, region) to support fairness analysis. These are coarse, corpus-derived categories used only in aggregate for representation measurement, not to profile or target individuals. Any real deployment should treat such attributes with care, obtain appropriate consent, and avoid individual-level inference.

\section*{Data Availability}

The OpinioBank hotel-review data used in our cross-domain evaluation is publicly available under the CC BY-NC-SA 4.0 license and was introduced by \citet{nayeem2025opiniorag}\footnote{\url{https://huggingface.co/datasets/tafseer-nayeem/OpinioBank}}; our use is for non-commercial research evaluation, consistent with the license. The e-commerce seller-forum corpus is proprietary and contains user-generated content that cannot be released for privacy and contractual reasons; we report aggregate statistics and anonymize all platform-specific terms and quoted content. All data is in English, and no additional preprocessing was applied beyond the chunking described in Appendix~\ref{app:reproducibility}. Opinion-extraction prompts and metric definitions are provided in the appendix to support reproduction on comparable corpora.

\section*{Use of AI Assistants}

We disclose the following use of AI assistants in preparing this work, separate from the LLMs studied as objects of research. Anthropic Claude Opus~4.6 assisted with writing the experiment code, Google Gemini supported the literature survey, and Claude Sonnet~4.5 provided grammatical \& fluency assistance for the manuscript. All research claims, analyses, and final text were written and reviewed by the authors.

\bibliography{custom}

\appendix
\section{Reproducibility and Experimental Settings}
\label{app:reproducibility}

\paragraph{Models.} Opinion extraction used three LLMs accessed through Amazon Bedrock in April 2026: Mistral Large, Qwen~235B, and Claude Sonnet~4.5. Response generation from the retrieved context used Anthropic Claude Sonnet~4.5. Document and query embeddings used Amazon Titan Text Embeddings~V2 (1024 dimensions).

\paragraph{Decoding.} All extraction and generation calls used temperature~0, top-$p$ at the Amazon Bedrock default, and the Bedrock default maximum output length of 4096~tokens. Because temperature is~0, outputs are effectively deterministic given fixed inputs, so all reported numbers come from a single run.

\paragraph{Retrieval.} Documents were indexed in an Amazon Bedrock Knowledge Base vector store using fixed-size chunking (4000~tokens, 15\% overlap). We report both semantic-only and hybrid (dense plus keyword) retrieval across depths $k \in \{10, 20, 40, 60\}$.

\paragraph{OpinioBank sampling.} We evaluate on a 3-property sample (Silverton Las Vegas, Ambassador Bangkok, Ala Moana Honolulu) drawn from the full 500-entity OpinioBank release. Entity and question counts reported in Section~\ref{sec:experiments} refer to this sample, not the full dataset.

\paragraph{Compute and cost.} All inference was API-hosted through Amazon Bedrock; the method involves no model training and no local GPU compute. Total inference cost across all extraction, generation, and ablation runs was approximately US\$2{,}500.

\paragraph{Statistical testing.} Significance is assessed with the Wilcoxon signed-rank test over paired questions; the confidence interval for human preference uses bootstrap resampling. Both were computed with the SciPy library (\texttt{scipy.stats}). Significance thresholds are stated in the relevant table captions.

\section{Opinion Extraction Model Variations}
\label{app:model_variations}

Table~\ref{tab:ablation_models} compares retrieval diversity when using different LLMs for opinion extraction. All configurations use hybrid search at $k{=}20$. Results show consistent improvements across models.

\begin{table}[H]
\centering
\small
\caption{Retrieval diversity improvements ($\Delta$ vs.\ Raw KB) across opinion extraction LLMs ($k{=}20$, hybrid).}
\label{tab:ablation_models}
\setlength{\tabcolsep}{4pt}
\begin{tabular}{lccc}
\toprule
\textbf{Metric} & \textbf{Mistral} & \textbf{Qwen} & \textbf{Sonnet} \\
\midrule
Sent-Int $\sigma$ & \textbf{+24.2\%} & \textbf{+27.4\%} & \textbf{+26.8\%}* \\
Biz Impact $\sigma$ & \textbf{+18.4\%} & \textbf{+12.4\%} & \textbf{+14.5\%} \\
Revenue Band Cov. & $-$17.5\% & $-$19.0\% & \textbf{$-$13.0\%}* \\
Tenure Cov. & $-$3.9\% & $-$3.6\% & \textbf{$-$6.9\%} \\
Evidence Cov. & \textbf{+9.4\%} & \textbf{+8.9\%} & +4.7\% \\
Emot.\ Markers & \textbf{+28.1\%} & \textbf{+20.0\%} & \textbf{+23.6\%}* \\
\bottomrule
\end{tabular}
\end{table}

\section{Retrieval Depth Analysis}
\label{app:depth}

Table~\ref{tab:depth} shows coverage and fairness metrics across retrieval depths.

\begin{table*}[t]
  \caption{Diversity metrics at different retrieval depths ($k$) using hybrid retrieval. \textbf{Bold}~=~$p{<}0.05$, *~=~$p{<}0.001$.}
  \label{tab:depth}
  \small
  \centering
  \begin{tabular}{llcccccccc}
    \toprule
    & & \textbf{Entity} & \multicolumn{2}{c}{\textbf{Scored Diversity ($\sigma$)}} & \multicolumn{3}{c}{\textbf{Categorical Coverage (\%)}} & \textbf{Emot.} \\
    \cmidrule(lr){4-5} \cmidrule(lr){6-8}
    $\boldsymbol{k}$ & \textbf{KB Config} & Match \% & Sent-Int & Biz Impact & Revenue Band & Tenure & Evidence & Marker \\
    \midrule
    \multirow{2}{*}{10} & Raw & 32.7 & 10.10 & 11.48 & 63.7 & 65.7 & 38.0 & 5.5 \\
    & Enriched & 48.5 & 13.51 & 12.73 & 51.8 & 59.2 & 40.1 & 7.0 \\
    \midrule
    \multirow{2}{*}{20} & Raw & 24.8 & 9.12 & 10.55 & 78.3 & 81.2 & 44.3 & 7.2 \\
    & Enriched & 35.4 & 11.56 & 12.08 & 68.1 & 75.6 & 46.4 & 8.9 \\
    \midrule
    \multirow{2}{*}{40} & Raw & 16.9 & 8.11 & 9.45 & 87.8 & 89.3 & 49.5 & 9.1 \\
    & Enriched & 24.6 & 9.68 & 10.88 & 83.6 & 85.1 & 53.1 & 11.0 \\
    \midrule
    \multirow{2}{*}{60} & Raw & 13.3 & 7.28 & 8.52 & 90.5 & 89.6 & 50.5 & 10.1 \\
    & Enriched & 19.0 & 8.52 & 9.78 & 89.0 & 87.5 & 53.6 & 12.2 \\
    \bottomrule
  \end{tabular}
\end{table*}

\section{Annotation Study Details}
\label{app:annotation}

The manual annotation comprised 129 forum posts labeled by 3 independent annotators. Raw agreement: 80.6\%. Gwet's AC1 = 0.845 (95\% CI [0.8, 0.9]). Fleiss' $\kappa$ = 0.23, which is low due to the kappa paradox under high prevalence (91\% positive votes); AC1 is the more appropriate metric under prevalence imbalance.

Annotators for both the labeling task and the blind pairwise preference evaluation were internal colleagues who participated as part of their normal work and consented to the use of their judgments; recruitment, compensation, consent, and ethics-review details are described in the Ethics Statement.

\paragraph{Pairwise preference instructions.} For the blind preference evaluation, each annotator was shown the two responses to a given question with their positions randomly assigned to slots A and B (so that neither the Raw KB nor the Opinion-Enriched KB was consistently placed). Annotators were told which system produced which response was hidden, and were asked a single question: which response, A or B, is preferred based on which better reflects a diverse and holistic range of opinions. No other scoring dimensions were collected.

\section{Entity-Split Ablation Details}
\label{app:split_ablation}

Tables~\ref{tab:split_ablation_sonnet} and~\ref{tab:split_ablation_mistral} disentangle per-entity splitting from opinion metadata enrichment across two extraction models.

\begin{table}[H]
\centering
\small
\caption{Entity-Split-Only ablation at $k{=}20$ (Sonnet 4.5). \textbf{Bold}~=~highest per metric.}
\label{tab:split_ablation_sonnet}
\setlength{\tabcolsep}{4pt}
\begin{tabular}{lccc}
\toprule
\textbf{KB Config} & \textbf{Sent-Int $\sigma$} & \textbf{Biz Impact} & \textbf{Emot Marker} \\
\midrule
Raw & 9.12 & 10.55 & 7.2 \\
Entity-Split Only & \textbf{13.1} & \textbf{12.92} & 7.8 \\
Full O-RAG & 11.56 & 12.08 & \textbf{8.9}* \\
\bottomrule
\end{tabular}
\end{table}

\begin{table}[H]
\centering
\small
\caption{Entity-Split-Only ablation at $k{=}20$ (Mistral Large). \textbf{Bold}~=~highest per metric, *~=~$p{<}0.001$ vs.\ Entity-Split.}
\label{tab:split_ablation_mistral}
\setlength{\tabcolsep}{4pt}
\begin{tabular}{lccc}
\toprule
\textbf{KB Config} & \textbf{Sent-Int $\sigma$} & \textbf{Biz Impact} & \textbf{Emot Marker} \\
\midrule
Raw & 6.53 & 9.24 & 3.2 \\
Entity-Split Only & 7.37 & 10.27 & 3.5 \\
Full O-RAG & \textbf{8.11}* & \textbf{10.94}* & \textbf{4.1}* \\
\bottomrule
\end{tabular}
\end{table}

\section{Qualitative Response Comparison}
\label{app:qualitative}

The following comparison illustrates how opinion enrichment affects generated responses. Key phrases are highlighted in \textcolor{red}{\textbf{red}}.

\noindent\textbf{Query:} \textit{``Give me a detailed overview of what sellers think about community management \& moderators specifically on seller forums?''}

Table~\ref{tab:qualitative_comparison} contrasts the Raw KB response against the Opinion-Enriched KB response.

\begin{table*}[t]
\centering
\small
\begin{tabular}{|p{0.47\textwidth}|p{0.47\textwidth}|}
\hline
\textbf{Raw Discussions KB Response} & \textbf{Opinion-Enriched KB Response} \\
\hline
Based on seller discussions in the forums, \textcolor{red}{\textbf{there are several perspectives}} on community management and moderators:

\textcolor{red}{\textbf{Requests for Moderator Intervention:}} Sellers frequently turn to moderators when they feel stuck in \textcolor{red}{\textbf{automated response loops}} or cannot get resolution through standard support channels. Multiple sellers specifically call out moderators in their posts, asking for help with account issues, policy violations, and reimbursement disputes.

\textcolor{red}{\textbf{Frustration with Support Systems:}} There's significant frustration with the overall support system that leads sellers to seek moderator help. Sellers report dealing with what they describe as a ``broken system.''

\textcolor{red}{\textbf{Moderator Responsiveness:}} Sellers appear to view moderators as a potential escalation path when standard channels fail. However, \textcolor{red}{\textbf{the search results don't show clear feedback}} on whether moderators consistently respond or resolve these escalated issues.

\textcolor{red}{\textbf{Community Guidelines Enforcement:}} At least one seller reported frustration with the moderation process for customer reviews, noting they had reported guideline violations but received no response after 17 days.
&
Sellers have \textcolor{red}{\textbf{mixed but generally positive}} views of community moderators, often seeing them as a \textcolor{red}{\textbf{critical escalation path}} when standard support channels fail. Multiple sellers report that \textcolor{red}{\textbf{moderators are the only ones who have successfully helped}} them resolve issues that automated systems could not address. Sellers turn to moderators when trapped in what they describe as \textcolor{red}{\textbf{``horrible loops''}} of unhelpful automated responses.

However, sellers \textcolor{red}{\textbf{express frustration}} with community management's responsiveness to guideline violations. One seller reported waiting 17 days without response. The reviews \textcolor{red}{\textbf{continued to negatively affect}} their listings and sales.

Sellers also \textcolor{red}{\textbf{request improvements}} to forum functionality. One seller suggested individual user accounts because the current setup causes irrelevant notifications. This seller indicated they \textcolor{red}{\textbf{``rarely ever post''}} specifically because of this issue.

When sellers reach out to moderators, it's typically \textcolor{red}{\textbf{as a last resort}} after exhausting other options. They view moderator intervention as \textcolor{red}{\textbf{necessary for protecting}} their accounts from false claims and removing malicious feedback. \\
\hline
\end{tabular}
\caption{Side-by-side comparison. The Raw KB organizes by \textit{action categories} and admits ``search results don't show clear feedback'' on outcomes. The Opinion-Enriched KB captures \textit{sentiment distribution} (mixed but positive), \textit{success attribution} (only ones who helped), \textit{behavioral explanations} (rarely post, last resort), and \textit{constructive suggestions} missing from Raw KB.}
\label{tab:qualitative_comparison}
\end{table*}

\section{Opinion Extraction Prompt}
\label{app:extraction_prompt}

We use an LLM with a structured output schema for opinion extraction. The prompt template is shown below (with platform-specific terms anonymized):

\begin{quote}
\small
\texttt{You are an expert at analyzing seller opinions from e-commerce platform content. Analyze the following post and extract structured opinion information for EACH entity (business topic/issue) mentioned.}

\texttt{POST CONTENT: \{content\}}\\
\texttt{PLATFORM: \{platform\_type\}}\\
\texttt{CATEGORY: \{platform\_category\}}\\
\texttt{\{entity\_registry\}}

\texttt{Extract the following information and respond ONLY with valid JSON:}
\begin{verbatim}
{"entities": [{
  "entity_name": "entity name",
  "confidence": "Low|Medium|High",
  "sentiment": "positive|negative|neutral|mixed",
  "sentiment_intensity": "Low|Medium|High",
  "emotional_markers": ["marker1", "marker2"],
  "evidence": "supporting evidence",
  "business_impact": "Low|Medium|High",
  "context_snippet": "relevant quote",
  "evidence_type": "personal|anecdotal|data"
}]}
\end{verbatim}
\end{quote}

Key extraction guidelines include:
\begin{itemize}
    \item \textbf{Entity matching}: If the entity matches one from the hierarchical registry, use that exact name; prefer granular (Tier 3) over general (Tier 1) entities
    \item \textbf{Per-entity extraction}: Each entity receives its own complete opinion structure; a single post may yield multiple entities with different sentiments
    \item \textbf{Evidence typing}: Classify as personal (first-hand experience), anecdotal (heard from others), or data (cites specific statistics)
    \item \textbf{Emotional markers}: Common markers include Frustration, Anger, Resignation, Adaptation, Hope, Satisfaction, Confusion, Anxiety
    \item \textbf{Business impact}: Consider financial mentions, operational disruption, and account risks when assigning impact levels
\end{itemize}

The model outputs a JSON array with one object per entity-opinion pair, enabling downstream per-entity document splitting.

\section{Question Type Ablation}
\label{app:question_type}

Table~\ref{tab:ablation_qtype} shows how question formulation affects diversity gains.

\begin{table}[H]
\centering
\caption{Sentiment-Intensity $\sigma$ by question type ($k{=}20$). Largest gains for Business Impact and Emotional queries; Experience queries benefit least.}
\label{tab:ablation_qtype}
\begin{tabular}{lccc}
\toprule
\textbf{Question Type} & \textbf{Raw} & \textbf{Enriched} & $\boldsymbol{\Delta}$ \\
\midrule
Detailed Overview & 9.14 & 11.80 & \textbf{+29.1\%} \\
Experience & 10.64 & 11.30 & \textbf{+6.2\%} \\
Business Impact & 8.56 & 11.89 & \textbf{+38.9\%} \\
Specific Examples & 9.02 & 11.14 & \textbf{+23.5\%} \\
Process & 7.82 & 10.44 & \textbf{+33.5\%} \\
Emotional & 9.56 & 12.82 & \textbf{+34.1\%} \\
\bottomrule
\end{tabular}
\end{table}

\section{Retrieval Method Comparison}
\label{app:retrieval_method}

Table~\ref{tab:ablation_retrieval} compares semantic-only and hybrid retrieval. Hybrid search amplifies opinion enrichment benefits (+26.8\%) over semantic-only (+21.4\%), as keyword matching surfaces entity-relevant documents.

\begin{table}[H]
\centering
\small
\caption{Semantic-only vs.\ hybrid retrieval at $k{=}20$. \textbf{Bold}~=~$p{<}0.05$, *~=~$p{<}0.001$.}
\label{tab:ablation_retrieval}
\setlength{\tabcolsep}{3pt}
\resizebox{\columnwidth}{!}{%
\begin{tabular}{lcccc}
\toprule
& \multicolumn{2}{c}{\textbf{Semantic}} & \multicolumn{2}{c}{\textbf{Hybrid}} \\
\cmidrule(lr){2-3} \cmidrule(lr){4-5}
\textbf{Metric} & Raw & Enr.\ ($\Delta$) & Raw & Enr.\ ($\Delta$) \\
\midrule
Sent-Int $\sigma$ & 8.93 & 10.84 (\textbf{+21.4\%*}) & 9.12 & 11.56 (\textbf{+26.8\%*}) \\
Biz Impact $\sigma$ & 10.47 & 11.63 (\textbf{+11.1\%}) & 10.55 & 12.08 (\textbf{+14.5\%}) \\
Revenue Band (\%) & 79.1 & 68.7 (\textbf{$-$13.1\%*}) & 78.3 & 68.1 (\textbf{$-$13.0\%*}) \\
Tenure (\%) & 81.5 & 76.8 (\textbf{$-$5.8\%}) & 81.2 & 75.6 (\textbf{$-$6.9\%}) \\
Evidence (\%) & 44.3 & 43.8 ($-$1.1\%) & 44.3 & 46.4 (+4.7\%) \\
Emot.\ Markers & 7.1 & 8.4 (\textbf{+18.3\%}) & 7.2 & 8.9 (\textbf{+23.6\%*}) \\
\bottomrule
\end{tabular}%
}
\end{table}

\section{Emergent Opinion Synthesis Example}
\label{app:responses}

The following is a response from ChatGPT (GPT-4o, January 2026) to the query: \textit{``What do small business owners think about the current tax policy?''}

\begin{quote}
\textit{Small business owners express mixed views on current tax policy. Many appreciate efforts toward permanence and certainty, particularly wanting key provisions from the 2017 Tax Cuts and Jobs Act made permanent to aid long-term planning. They support targeted deductions and pass-through relief, especially proposals allowing deduction of the first \$25,000 of qualifying business income. However, concerns persist: many feel previous tax cuts primarily benefited larger corporations rather than small businesses. Common complaints include tax code complexity, high compliance costs, and perceived unfairness in how large firms can leverage specialists and loopholes. Small business owners consistently advocate for simplified reporting, scaled incentives that actually help smaller firms, and measures ensuring ``fair play'' against large corporation advantages.}
\end{quote}

This response sounds balanced, but without transparency into retrieval, users cannot determine whether it reflects genuine diversity in retrieved sources. Or the model's learned tendency to present information evenhandedly regardless of input.

\section{Biased Leadership Response Example}
\label{app:biased_leadership}

The following illustrates how traditional RAG metrics can assign high scores to biased outputs. In response to the query \textit{``What leadership qualities are most important for startup CEOs?''}, a system might generate:

\begin{quote}
\textit{Successful startup CEOs demonstrate strong analytical thinking, decisive action under pressure, and natural assertiveness in competitive markets. Research shows that effective leaders maintain composure during high-stakes negotiations, exhibit confidence in pitching to investors, and demonstrate the aggressive persistence needed to disrupt established industries. The most successful CEOs are typically those who embrace calculated risks, maintain unwavering conviction in their vision, and possess the competitive drive necessary to outperform rivals.}
\end{quote}

While scoring excellently across standard metrics, this response amplifies gender bias by emphasizing traditionally masculine-coded traits while omitting empathetic or inclusive leadership qualities.


\section{Entity-Matched Document Analysis}
\label{app:entity_matched}

Table~\ref{tab:ablation_matching} analyzes diversity metrics on only the subset of retrieved documents that mention the queried entity, explaining the diversity-fairness trade-off discussed in Section~\ref{sec:experiments}.

\begin{table}[H]
\centering
\small
\caption{Diversity metrics on entity-matched documents only ($k{=}20$, forums dataset). On entity-matched documents, Opinion-Enriched KB \textit{improves} demographic coverage- the opposite of the overall result.}
\label{tab:ablation_matching}
\begin{tabular}{lccc}
\toprule
\textbf{Metric} & \textbf{Raw} & \textbf{Enr.} & $\boldsymbol{\Delta}$ \\
\midrule
Matched Docs (of 960) & 238 & 340 & \textbf{+42.9\%}* \\
\midrule
Sent-Int $\sigma$ & 14.29 & 16.57 & +16.0\% \\
Biz Impact $\sigma$ & 4.45 & 5.47 & \textbf{+22.9\%} \\
Revenue Band Cov.\ (\%) & 42.9 & 47.0 & \textbf{+9.6\%} \\
Tenure Cov.\ (\%) & 40.5 & 53.3 & \textbf{+31.6\%}* \\
Evidence Cov.\ (\%) & 44.3 & 46.4 & +4.7\% \\
Emot.\ Markers & 7.2 & 8.9 & \textbf{+23.6\%}* \\
\bottomrule
\end{tabular}
\end{table}

\section{Extended Benchmark Analysis}
\label{app:benchmarks}

Table~\ref{tab:rag_benchmarks_full} provides a comprehensive analysis of 36 major evaluation datasets used in RAG research from 2013--2025.

\begin{table*}[t]
\centering
\small
\setlength{\tabcolsep}{4pt}
\begin{tabular}{|l|c|l|l|c|l|}
\hline
\textbf{Dataset} & \textbf{Year} & \textbf{Task Category} & \textbf{Evaluation Focus} & \textbf{Opinion} & \textbf{Reference} \\
\hline
\multicolumn{6}{|c|}{\textbf{Single-hop Question Answering}} \\
\hline
WebQuestions & 2013 & Knowledge Base QA & EM, F1 & No & \cite{berant2013semantic} \\
MS MARCO & 2016 & Machine Reading Comprehension & ROUGE-L, BLEU & No & \cite{nguyen2016ms} \\
TriviaQA & 2017 & Reading Comprehension & EM, F1 & No & \cite{joshi2017triviaqa} \\
SQuAD 2.0 & 2018 & Reading Comprehension & EM, F1 & No & \cite{rajpurkar2016squad} \\
Natural Questions & 2019 & Open-domain QA & EM, Token F1 & No & \cite{kwiatkowski2019natural} \\
PopQA & 2023 & Entity-centric QA & Accuracy & No & \cite{popqa2023} \\
\hline
\multicolumn{6}{|c|}{\textbf{Multi-hop Question Answering}} \\
\hline
HotpotQA & 2018 & Multi-hop Reasoning & EM, F1, Support F1 & No & \cite{yang2018hotpotqa} \\
2WikiMultiHopQA & 2020 & Multi-hop Reasoning & EM, F1 & No & \cite{wikimultihopqa2021} \\
MuSiQue & 2022 & Multi-hop Reasoning & EM, F1 & No & \cite{musique2022} \\
\hline
\multicolumn{6}{|c|}{\textbf{Long-form Question Answering}} \\
\hline
NarrativeQA & 2018 & Long-form QA & ROUGE, BLEU & No & \cite{narrativeqa2018} \\
ELI5 & 2019 & Long-form QA & ROUGE-L & No & \cite{eli5_2019} \\
QMSum & 2021 & Meeting Summarization & ROUGE & No & \cite{qmsum2021} \\
ASQA & 2022 & Ambiguous QA & ROUGE, QA-based & No & \cite{asqa2022} \\
\hline
\multicolumn{6}{|c|}{\textbf{Domain-specific Question Answering}} \\
\hline
COVID-QA & 2020 & Medical Domain & EM, F1 & No & \cite{covidqa2020} \\
Qasper & 2021 & Scientific Papers & EM, F1 & No & \cite{qasper2021} \\
CMB & 2024 & Chinese Medical & Accuracy & No & \cite{cmb2023} \\
\hline
\multicolumn{6}{|c|}{\textbf{Multi-choice Question Answering}} \\
\hline
ARC & 2018 & Science Questions & Accuracy & No & \cite{arc2018} \\
CommonsenseQA & 2019 & Commonsense Reasoning & Accuracy & No & \cite{commonsenseqa2019} \\
MMLU & 2021 & World Knowledge & Accuracy & No & \cite{mmlu2020} \\
QuALITY & 2022 & Reading Comprehension & Accuracy & No & \cite{quality2021} \\
\hline
\multicolumn{6}{|c|}{\textbf{Information Retrieval \& Knowledge Tasks}} \\
\hline
KILT & 2021 & Knowledge-Intensive Tasks & EM, F1, ROUGE-L & No & \cite{kilt2020} \\
BEIR & 2021 & Information Retrieval & nDCG, MAP, Recall & No & \cite{beir2021} \\
\hline
\multicolumn{6}{|c|}{\textbf{Fact Verification \& Robustness}} \\
\hline
FEVER & 2018 & Fact Verification & Accuracy, FEVER Score & No & \cite{thorne2018fever} \\
PubHealth & 2020 & Health Fact-checking & Accuracy & No & \cite{pubhealth2020} \\
TruthfulQA & 2022 & Truthfulness Evaluation & \% True, \% Informative & Partial & \cite{truthfulqa2021} \\
NoMIRACL & 2024 & Multilingual Robustness & Hallucination Rate, Error Rate & No & \cite{nomiracl2023} \\
CRAG & 2024 & Comprehensive QA & Accuracy, Hallucination & No & \cite{yang2024crag} \\
RAGuard & 2025 & Robustness Testing & Resistance to Misinfo & No & \cite{zeng2025worse} \\
CONFLICTS & 2025 & Conflict Resolution & Appropriateness Score & No & \cite{cattan2025dragged} \\
\hline
\multicolumn{6}{|c|}{\textbf{Generation \& Attribution}} \\
\hline
ALCE & 2023 & Citation Generation & Citation Quality, Fluency & No & \cite{alce2023} \\
FActScore & 2023 & Factual Precision & Atomic Fact Accuracy & No & \cite{factscore2023} \\
AttrScore & 2023 & Attribution Quality & Attribution Accuracy & No & \cite{attrscore2023} \\
RAGBench & 2024 & Industry RAG & TRACe Metrics & No & \cite{friel2024ragbench} \\
OpinioRAG & 2025 & Opinion Summarization & Sentiment Alignment & Yes & \cite{nayeem2025opiniorag} \\
\hline
\multicolumn{6}{|c|}{\textbf{Perspective-Aware Retrieval}} \\
\hline
PIR & 2024 & Perspective Retrieval & p-Recall, Bias & Partial & \cite{zhao2024pir} \\
BERDS & 2025 & Diverse Perspective Retrieval & Perspective Coverage & Partial & \cite{chen2025berds} \\
\hline
\end{tabular}
\caption{Complete analysis of 36 major RAG/LLM evaluation datasets (2013--2025). Only OpinioRAG addresses opinion synthesis; PIR and BERDS evaluate perspective-aware retrieval but not distributional fidelity.}
\label{tab:rag_benchmarks_full}
\end{table*}

\end{document}